\documentclass[10pt]{article}

\usepackage[margin=1in]{geometry}
\usepackage[T1]{fontenc}
\usepackage[utf8]{inputenc}
\usepackage{newtxtext}
\usepackage{newtxmath}
\usepackage{graphicx}
\usepackage{booktabs}
\usepackage{amsmath}
\usepackage{array}
\usepackage{xcolor}
\usepackage[hidelinks]{hyperref}
\usepackage{caption}
\usepackage{enumitem}
\usepackage{authblk}
\usepackage{multirow}
\usepackage{makecell}
\usepackage{float}
\usepackage{microtype}

\graphicspath{{./}}
\setlist{nosep,leftmargin=1.8em}
\title{Synthetic Point Cloud Generation and 3D Object Detection\\
Based on Per-Axis Quantile Calibration and VoxelRCNN}
\newcommand{\teamname}{\textbf{Team Name:} loopy}

\author[1]{Antong Deng}
\author[1]{Yuxin Zhang}
\author[1]{Zhongxiang Xing}
\author[1]{Zhenzhen Lin}
\author[1]{Jiong Mu}

\affil[1]{%
College of Information Engineering, Sichuan Agricultural University,
Ya'an, Sichuan 625014, China\\
\small

\textbf{E-mail:}
2937807858@qq.com
}
\date{}

\begin{document}
\maketitle
\begin{center}
\teamname
\end{center}
\vspace{0.4em}

\begin{abstract}
This report presents a complete technical solution for the UrbanTwin V2X-Real Sim2Real LiDAR Challenge, covering both synthetic point cloud generation and 3D object detection. For data generation, we adopt per-axis quantile calibration. Public V2X-Real training data are used to construct the real reference distribution, and independent one-dimensional quantile mappings are estimated for the $x$, $y$, and $z$ axes of the base synthetic point clouds. The calibrated point clouds are then resampled using a fixed random seed and clipped to the valid spatial range, producing 50 frames with 2,000 float32 XYZ points per frame. For detection, we use a two-stage Voxel R-CNN (VoxelRCNN) detector trained with RangeLDM-enhanced synthetic point-cloud data, point clouds downsampled to 50k points per frame, and reconstructed pedestrian data. Training is performed on four NVIDIA GeForce RTX 4090 GPUs with a global batch size of 16 for 40 epochs. Final testing uses direct single-model inference with the original VoxelRCNN model. On the official evaluation server, the normalized Realism score is 0.7905 and the normalized Detection score is 0.5583, with mAP@0.5 of 0.3908, BEV mAP@0.5 of 0.4219, and mAP@0.7 of 0.1047. According to the official weighting rule, the final combined score is $0.6\times0.5583+0.4\times0.7905=0.6512$. This report documents the complete implementation details of data construction, model training, inference, submission formatting, and official evaluation.
\end{abstract}


\section{Introduction}

The UrbanTwin V2X-Real Sim2Real LiDAR Challenge jointly evaluates the realism of synthetic point clouds and the generalization capability of 3D object detectors on the real target domain. The realism task requires a fixed number of synthetic LiDAR frames and evaluates their consistency with real data using geometric and distributional metrics. The detection task requires a detector trained only with allowed training data to predict 3D bounding boxes for vehicles, pedestrians, and trucks in the target-domain test set.

Our solution organizes the two tasks into independent but data-consistent branches. The Realism branch performs statistical calibration of the base synthetic point clouds, resampling, and final submission generation. The Detection branch constructs multi-source training data, trains VoxelRCNN, and performs inference on the test set. The two branches finally produce the synthetic point cloud files and the detection prediction file required by the official server.

The core of the synthetic point cloud generation procedure is per-axis quantile calibration. We first estimate one-dimensional empirical distributions from a large pool of base synthetic points and public real reference points, then construct monotonic source-to-target mappings using 2,001 quantile levels. For every synthetic point, the three coordinate components are calibrated independently using the corresponding axis-specific mapping. After the coordinate transformation, fixed-seed resampling and spatial clipping are applied to produce submission point clouds with a unified format.

For 3D object detection, we adopt the two-stage Voxel R-CNN architecture. The training data consist of RangeLDM-enhanced synthetic point-cloud data, point clouds downsampled to 50k points per frame, and reconstructed pedestrian data. VoxelRCNN first generates 3D proposals from voxel features and then refines each proposal with RoI-level voxel features, producing the final category scores and 3D box regression outputs.

The remainder of this report describes the overall framework, synthetic point cloud generation method, VoxelRCNN detector, engineering implementation, evaluation metrics, and official results.

\section{Overall Framework}

\begin{figure*}[t]
    \centering
    \includegraphics[width=\textwidth]{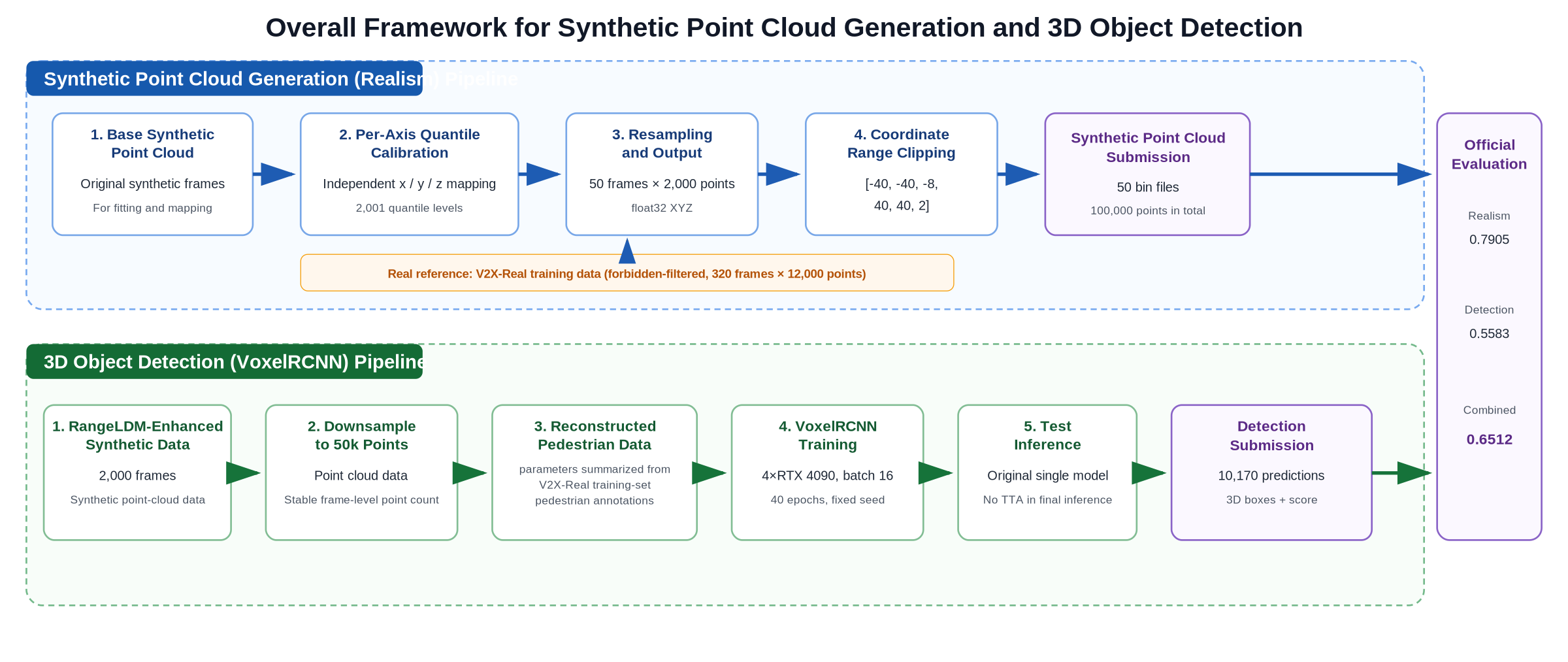}
    \caption{Overall framework for synthetic point cloud generation and 3D object detection. The Realism branch applies per-axis quantile calibration, resampling, and spatial clipping to the base synthetic point clouds. The Detection branch trains VoxelRCNN with RangeLDM-enhanced synthetic point-cloud data, point clouds downsampled to 50k points per frame, and reconstructed pedestrian data, followed by direct single-model inference on the test set.}
    \label{fig:framework}
\end{figure*}

Figure~\ref{fig:framework} presents the complete workflow. The upper branch is the Realism pipeline and the lower branch is the Detection pipeline. The two branches are processed independently during data generation and model training, and finally produce separate point cloud and detection submissions for official evaluation.

\noindent\fcolorbox{blue!65!black}{blue!3}{%
\parbox{0.965\linewidth}{\textbf{TTA exploration (tried but not used).}
We evaluated flip-4 test-time augmentation using the original input, \texttt{flip\_x}, \texttt{flip\_y}, and \texttt{flip\_xy} with prediction fusion.
\textbf{Local val50 validation: mAP 0.535 $\rightarrow$ 0.397, Truck AP 0.788 $\rightarrow$ 0.349; therefore, flip-4 TTA was discarded.}
These values are local validation results only. The final official submission uses direct single-model inference with the original VoxelRCNN model and no multi-variant fusion.}}
\vspace{0.6em}

\subsection{Realism Branch}

The Realism branch contains five main steps: loading the base synthetic data, pooling synthetic point clouds, constructing the real reference pool, applying per-axis quantile calibration, and performing final resampling and spatial clipping. The source and real reference point clouds are used only to estimate the statistical mapping. The final output strictly follows the required number of frames, point format, and coordinate range.

The real reference pool is built from public V2X-Real training frames after forbidden-data filtering. A fixed random seed is used to sample 320 frames, with 12,000 points pooled from each frame. The base synthetic frames come from the same synthetic source used for final submission and are aggregated in the same coordinate definition to form the source pool.

\subsection{Detection Branch}

The Detection branch starts from RangeLDM-enhanced synthetic point-cloud data. Each frame is downsampled to 50,000 points to standardize the frame-level point count. Reconstructed pedestrian data are generated using parameters summarized from V2X-Real training-set pedestrian annotations. The resulting data are used to train VoxelRCNN. At test time, the trained checkpoint is loaded directly, predictions are generated for all target test frames, and the final boxes are converted to the required LiDAR coordinate format.

\section{Synthetic Point Cloud Generation}

\subsection{Data Sources and Output Specification}

Two data sources are used during synthetic point cloud generation: base synthetic point clouds and real reference point clouds. The base synthetic point clouds serve as the source distribution to be calibrated, while the real point clouds are used only to estimate the target distribution. After forbidden-data filtering, 320 real training frames are pooled with a fixed random seed, and 12,000 valid points are sampled from each frame.

The final synthetic submission is generated with the following fixed specification:
\begin{itemize}
    \item Number of frames: 50;
    \item Points per frame: 2,000;
    \item Data type: float32;
    \item Point attributes: XYZ;
    \item Data size per frame: $2{,}000\times3\times4=24{,}000$ bytes;
    \item Total number of points: 100,000;
    \item Coordinate range: $[-40,-40,-8,40,40,2]$.
\end{itemize}

The coordinate range is expressed as $(x_{\min},y_{\min},z_{\min},x_{\max},y_{\max},z_{\max})$. Points outside the valid range are clipped before output.

\subsection{Per-Axis Quantile Calibration}

Let the base synthetic point cloud set be $\mathcal{S}$ and the real reference point cloud set be $\mathcal{T}$. For each coordinate axis $a\in\{x,y,z\}$, all coordinate values are collected as
\begin{equation}
S_a=\{p_a\mid p\in\mathcal{S}\},\qquad
T_a=\{q_a\mid q\in\mathcal{T}\}.
\end{equation}

We define 2,001 uniformly spaced quantile levels:
\begin{equation}
\mathcal{P}=\{p_i\}_{i=0}^{2000},\qquad p_i=\frac{i}{2000}.
\end{equation}

For each axis, source and target quantile knots are computed as
\begin{equation}
Q^{S}_{a,i}=\operatorname{Quantile}(S_a,p_i),\qquad
Q^{T}_{a,i}=\operatorname{Quantile}(T_a,p_i).
\end{equation}

For a synthetic point $\mathbf{x}=(x,y,z)$ and one coordinate $x_a$, we first locate the adjacent source interval $[Q^{S}_{a,i},Q^{S}_{a,i+1}]$ and then map the value to the corresponding target interval by piecewise linear interpolation:
\begin{equation}
\hat{x}_a=
Q^{T}_{a,i}+\lambda\left(Q^{T}_{a,i+1}-Q^{T}_{a,i}\right),
\end{equation}
where
\begin{equation}
\lambda=\frac{x_a-Q^{S}_{a,i}}{Q^{S}_{a,i+1}-Q^{S}_{a,i}+\epsilon}.
\end{equation}

Applying this operation independently to the $x$, $y$, and $z$ coordinates gives the calibrated point
\begin{equation}
\hat{\mathbf{x}}=(\hat{x},\hat{y},\hat{z}).
\end{equation}

In the implementation, the same operation is performed using one-dimensional piecewise linear interpolation:
\begin{equation}
\texttt{output[:,a]}=\texttt{interp}(\texttt{points[:,a]},Q^{S}_a,Q^{T}_a).
\end{equation}

\subsection{Resampling and File Generation}

After per-axis calibration, each submission frame is resampled with a fixed random seed so that the number of points is exactly 2,000. Frames with more than the target number of points are sampled without replacement. If fewer than 2,000 valid points remain after processing, additional points are sampled from the valid set according to the deterministic completion rule. The final coordinates are converted to float32 and written as contiguous XYZ triplets.

To ensure reproducibility, the entire data-processing pipeline uses fixed random seeds. Before writing the final submission, we check the number of files, number of points per frame, data type, non-finite values, coordinate range, and total file size. All submission frames are generated by the same processing script to guarantee a consistent output format.

\section{VoxelRCNN 3D Object Detection}

\subsection{Model Architecture}

We adopt the two-stage Voxel R-CNN detector. The input point cloud is first voxelized so that the irregular point set is converted into a sparse 3D voxel representation. Sparse voxel features are processed by a 3D backbone network to extract multi-scale representations and are subsequently projected to bird's-eye-view features. The first-stage detection head generates 3D proposals from the shared features. The second stage aggregates RoI features from the 3D voxel representation and refines the proposal classification and box regression outputs.

Each final prediction is represented as
\begin{equation}
\mathbf{b}=(x,y,z,d_x,d_y,d_z,\theta,c,s),
\end{equation}
where $(x,y,z)$ is the 3D center, $(d_x,d_y,d_z)$ is the box size, $\theta$ is the heading angle, $c$ is the category, and $s$ is the confidence score. All predictions are converted to the required LiDAR coordinate system before submission.

\subsection{RangeLDM-Enhanced Synthetic Point-Cloud Data}

The RangeLDM-enhanced synthetic point-cloud set contains 2,000 frames and is used as a primary synthetic training source for VoxelRCNN. The corresponding 3D object annotations are retained, and all frames are checked for coordinate consistency and class-definition consistency before training.

\subsection{Point Clouds Downsampled to 50k Points}

To stabilize the input size during training, each point cloud frame is downsampled to approximately 50,000 points. This preprocessing step reduces frame-to-frame variation in the original point count and produces more consistent training batches. For frames with more than 50,000 points, random downsampling is performed according to a fixed processing rule while preserving the associated 3D annotations.

\subsection{Reconstructed Pedestrian Data}

Reconstructed pedestrian data are generated using parameters summarized from V2X-Real training-set pedestrian annotations. These summarized parameters provide the basis for adapting synthetic pedestrian instances during detector training.

\subsection{Training Configuration}

The detector is trained using the GBlobs implementation of VoxelRCNN. The main training configuration is summarized in Table~\ref{tab:train}.

\begin{table}[H]
\centering
\footnotesize
\setlength{\tabcolsep}{8pt}
\begin{tabular}{lc}
\toprule
\textbf{Configuration} & \textbf{Setting} \\
\midrule
Detector & VoxelRCNN (GBlobs fork) \\
Training hardware & 4$\times$NVIDIA GeForce RTX 4090 \\
Global batch size & 16 (4 / GPU) \\
Training epochs & 40 \\
DataLoader workers & 8 \\
Random seed & Fixed \\
Training data & RangeLDM-enhanced synthetic point-cloud data + point clouds downsampled to 50k points + reconstructed pedestrian data \\
Inference & Direct single-model inference \\
\bottomrule
\end{tabular}
\caption{VoxelRCNN training and inference configuration.}
\label{tab:train}
\end{table}

After training, the final checkpoint is loaded directly for test-set inference. No additional geometric transformation is applied during inference, and no multi-variant fusion is used. Network outputs are decoded and converted to the required LiDAR coordinate system before generating the final prediction file.

\subsection{Local TTA Validation and Final Inference Strategy}

Before selecting the final inference strategy, we evaluated flip-4 TTA on the local val50 validation set. The four input variants were the original input, \texttt{flip\_x}, \texttt{flip\_y}, and \texttt{flip\_xy}, followed by prediction fusion. On the local val50 validation set, the original single-model VoxelRCNN achieved an mAP of 0.535, whereas flip-4 TTA reduced mAP to 0.397. Truck AP decreased from 0.788 to 0.349. Based on these local validation results, flip-4 TTA was discarded, and the official test set was processed using direct single-model inference with the original VoxelRCNN model. These values are reported only to document the local inference-strategy selection and are not official online evaluation results.

\section{Engineering Implementation and Submission Format}

\subsection{Point Cloud Processing}

Per-axis quantile calibration is implemented using Python and NumPy. The source and target point clouds are first loaded and concatenated, and 2,001 quantile knots are computed independently along all three axes. Because the estimated mappings remain fixed after fitting, the same set of knots is reused for all submission frames rather than recomputed frame by frame.

For each input point cloud, the processing sequence is: finite-value validation, per-axis mapping, valid-range clipping, fixed-seed sampling, float32 conversion, and binary writing. The output files contain only the XYZ coordinates and do not include semantic labels or intensity values.

\subsection{Detection Implementation}

VoxelRCNN uses the two-stage implementation in the GBlobs fork. Basic point cloud preprocessing is performed in the data loader, after which the data enter the voxelization module. The inference stage maintains the same coordinate definition as the training pipeline, and 3D boxes are directly exported in the LiDAR coordinate system after post-processing.

The final test inference produces 10,170 predictions in total: 4,133 vehicle predictions, 3,752 pedestrian predictions, and 2,285 truck predictions. Each prediction contains the category, confidence score, and 3D bounding-box geometry.

\subsection{Submission Validation}

Before submission, automated checks are applied independently to the Realism and Detection outputs. For Realism, we verify that all 50 point cloud files are present, each frame contains exactly 2,000 points, the data type is float32 XYZ, all coordinates lie within the valid range, and the file size satisfies the submission constraints. For Detection, we verify that every test frame has a prediction record, class names are valid, all 3D box parameters are finite, and all predictions use the required LiDAR coordinate system.

\section{Evaluation Metrics}

\subsection{Realism Metrics}

Realism evaluation includes CD, MMD, EMD, and FPD. All four raw metrics are lower-is-better. The server normalizes each metric and then combines them into the final normalized Realism score.

Chamfer Distance (CD) measures bidirectional nearest-neighbor distance between point sets. Maximum Mean Discrepancy (MMD) measures the difference between sample distributions in kernel space. Earth Mover's Distance (EMD) estimates the optimal transport cost between two point sets. Fr\'echet Point-cloud Distance (FPD) compares the mean and covariance of point-cloud feature distributions. Together, these metrics characterize geometric and statistical similarity between the submitted synthetic point clouds and the real reference data.

\subsection{Detection Metrics}

Detection evaluation includes mAP@0.5, BEV mAP@0.5, mAP@0.7, and category-specific AP for vehicle, pedestrian, and truck. mAP@0.5 measures overall 3D detection quality under a relatively permissive IoU threshold, BEV mAP@0.5 evaluates overlap in bird's-eye view, and mAP@0.7 uses a stricter threshold to assess 3D localization accuracy. The official protocol further combines the detection metrics into the normalized Detection score.

\section{Official Evaluation Results}

\subsection{Realism Results}

Table~\ref{tab:realism} reports the official Realism results. The final normalized Realism score is 0.7905. The four raw distance metrics are CD 6.90, MMD $1.25\times10^{-2}$, EMD 2.26, and FPD 9.34.

\begin{table}[H]
\centering
\footnotesize
\setlength{\tabcolsep}{12pt}
\begin{tabular}{lc}
\toprule
\textbf{Metric} & \textbf{Official Result} \\
\midrule
Normalized Realism & \textbf{0.7905} \\
CD & 6.90 \\
MMD & $1.25\times10^{-2}$ \\
EMD & 2.26 \\
FPD & 9.34 \\
\bottomrule
\end{tabular}
\caption{Official Realism evaluation results for the synthetic point cloud submission.}
\label{tab:realism}
\end{table}

These results correspond to the final submission of 50 frames with 2,000 points per frame, for a total of 100,000 float32 XYZ points.

\subsection{Detection Results}

The official Detection results are summarized in Table~\ref{tab:detection}. The normalized Detection score is 0.5583, mAP@0.5 is 0.3908, BEV mAP@0.5 is 0.4219, and mAP@0.7 is 0.1047. Category-specific AP is 0.5672 for vehicle, 0.0825 for pedestrian, and 0.5227 for truck.

\begin{table}[H]
\centering
\footnotesize
\setlength{\tabcolsep}{10pt}
\begin{tabular}{lc}
\toprule
\textbf{Metric} & \textbf{Official Result} \\
\midrule
Normalized Detection & \textbf{0.5583} \\
mAP@0.5 & 0.3908 \\
BEV mAP@0.5 & 0.4219 \\
mAP@0.7 & 0.1047 \\
AP$_{vehicle}$ & 0.5672 \\
AP$_{pedestrian}$ & 0.0825 \\
AP$_{truck}$ & 0.5227 \\
\midrule
Number of predictions & 10,170 \\
Number of GT instances & 1,547 \\
Prediction / GT ratio & 6.6$\times$ \\
\bottomrule
\end{tabular}
\caption{Official VoxelRCNN Detection evaluation results.}
\label{tab:detection}
\end{table}

The class-wise prediction counts are 4,133 vehicles, 3,752 pedestrians, and 2,285 trucks, giving 10,170 predicted instances in total. All predictions are produced directly by the same VoxelRCNN checkpoint.

\subsection{Combined Score}

The official combined score uses a Detection weight of 0.6 and a Realism weight of 0.4:
\begin{equation}
\mathrm{Combined}=0.6\times0.5583+0.4\times0.7905=0.65118\approx\mathbf{0.6512}.
\end{equation}
Therefore, the final official combined score of the proposed solution is \textbf{0.6512}.

\begin{table}[H]
\centering
\footnotesize
\setlength{\tabcolsep}{10pt}
\begin{tabular}{lcc}
\toprule
\textbf{Item} & \textbf{Weight / Formula} & \textbf{Result} \\
\midrule
Detection & 0.6 & 0.5583 \\
Realism & 0.4 & 0.7905 \\
Combined & $0.6\times0.5583+0.4\times0.7905$ & \textbf{0.6512} \\
\bottomrule
\end{tabular}
\caption{Official combined score.}
\label{tab:combined}
\end{table}

\section{Characteristics of the Solution}

The synthetic point cloud generation branch establishes a deterministic statistical mapping through fixed quantile levels. Quantile knots are computed once over the complete source and target pools and are then reused for every submission frame, resulting in a clear and reproducible input-output relationship. All 50 output frames follow exactly the same point count, data type, and spatial range, which supports consistent server-side sampling and metric computation.

The detection branch separates data processing from detector optimization. RangeLDM-enhanced synthetic point-cloud data provide the base training samples, the 50k-point downsampling step standardizes the frame-level point count, and reconstructed pedestrian data use parameters summarized from V2X-Real training-set pedestrian annotations. The processed data are then used consistently to train VoxelRCNN. The two-stage architecture introduces an RoI refinement stage after proposal generation, allowing classification and 3D box regression to be optimized a second time at the proposal level.

From an engineering perspective, the Realism branch requires only NumPy-level statistical computation and file processing. Its main computational cost is concentrated in the one-time quantile estimation and batch coordinate interpolation. The Detection branch is dominated by VoxelRCNN training on four RTX 4090 GPUs. Final inference and submission generation are fully scriptable and can be reproduced in a unified server environment.

\section{Conclusion}

This report presents a complete synthetic point cloud generation and 3D object detection solution. The point cloud generation branch applies per-axis quantile calibration using 2,001 quantile levels independently on the $x$, $y$, and $z$ coordinates of the base synthetic point clouds, followed by fixed-seed resampling and spatial clipping to generate 50 submission frames with 2,000 points each. The detection branch trains VoxelRCNN using RangeLDM-enhanced synthetic point-cloud data, point clouds downsampled to 50k points per frame, and reconstructed pedestrian data whose parameters are summarized from V2X-Real training-set pedestrian annotations, on four NVIDIA GeForce RTX 4090 GPUs. Based on local val50 validation of flip-4 TTA, the final inference strategy uses the original VoxelRCNN single model without TTA.

On the official evaluation server, the normalized Realism score is 0.7905, the normalized Detection score is 0.5583, and mAP@0.5 is 0.3908. With the official weights of 0.6 for Detection and 0.4 for Realism, the final Combined score is \textbf{0.6512}. The data-processing procedure, model training configuration, inference strategy, and submission format described in this report correspond to the final submitted solution and can be used as a technical record for reproduction.


\begin{thebibliography}{9}
\bibitem{voxelrcnn}
J. Deng, S. Shi, P. Li, W. Zhou, Y. Zhang, and H. Li,
``Voxel R-CNN: Towards High Performance Voxel-based 3D Object Detection,''
\emph{Proceedings of the AAAI Conference on Artificial Intelligence}, 2021.

\bibitem{challenge}
UrbanTwin V2X-Real Sim2Real LiDAR Challenge, official challenge evaluation protocol and submission specification, 2026.
\end{thebibliography}
\end{document}